\documentclass{article}
\usepackage{spconf,amsmath,graphicx}

\usepackage{enumitem}
\setlist{nosep, leftmargin=14pt}

\usepackage{mwe} 
\usepackage{booktabs}
\usepackage{multirow}
\usepackage{xcolor,colortbl}

\usepackage{cite}
\usepackage{amsmath,amssymb,amsfonts,bbm}
\usepackage{algorithmic}
\usepackage{graphicx}
\usepackage{textcomp}
\usepackage{xcolor}
\usepackage{enumitem}
\usepackage{booktabs}
\usepackage{multirow}
\usepackage{xcolor,colortbl}
\usepackage{setspace}
\usepackage{url} 

\def\BibTeX{{\rm B\kern-.05em{\sc i\kern-.025em b}\kern-.08em
    T\kern-.1667em\lower.7ex\hbox{E}\kern-.125emX}}
\DeclareMathOperator*{\topk}{Topk}

\title{Multi-source-free Domain Adaptation via Uncertainty-aware Adaptive Distillation}
\name{Yaxuan Song, Jianan Fan, Dongnan Liu, Weidong Cai}
\address{School of Computer Science, University of Sydney, Australia}
\begin{document}
%
\maketitle
\begin{abstract}
Source-free domain adaptation (SFDA) alleviates the domain discrepancy among data obtained from domains without accessing the data for the awareness of data privacy.
However, existing conventional SFDA methods face inherent limitations in medical contexts, where medical data are typically collected from multiple institutions using various equipment.
To address this problem, we propose a simple yet effective method, named \textit{\textbf{U}ncertainty-aware \textbf{A}daptive \textbf{D}istillation} (UAD) for the multi-source-free unsupervised domain adaptation (MSFDA) setting. 
UAD aims to perform well-calibrated knowledge distillation from (i) model level to deliver coordinated and reliable base model initialisation and (ii) instance level via model adaptation guided by high-quality pseudo-labels, thereby obtaining a high-performance target domain model.
To verify its general applicability, we evaluate UAD on two image-based diagnosis benchmarks among two multi-centre datasets, where our method shows a significant performance gain compared with existing works. 
The code is available at \url{https://github.com/YXSong000/UAD}.
\end{abstract}
\begin{keywords}
Unsupervised Domain Adaptation, Multi-source-free, Uncertainty-ware
\end{keywords}
\section{Introduction}
\label{sec:intro}

Unsupervised domain adaptation (UDA) is a promising streamline of works to compensate for the distributional discrepancy\cite{Fan_2023_ICCV}.
It seeks to utilise existing transferable knowledge from labelled data drawn from one or more source domains to recognise unlabelled data in the target domain\cite{ben-david_theory_2010}. 
UDA has shown great success in a broad spectrum of downstream applications, including classification \cite{pmlr-v37-ganin15} \cite{liang2020we} \cite{li2022domain}, segmentation \cite{liu2020unsupervised} \cite{wang2022cris} \cite{fan2024learning} and object detection \cite{hsu2020progressive} \cite{Liu2022DecomposeTA} by mitigating this domain shift. 

Despite its great promises in general visual perception tasks, existing UDA approaches inherently fall short in medical scenarios where additional regulations on data sharing restrictions.
To address the problems on medical images, source-free DA methods \cite{liang2020we} have been developed, providing the pre-trained source model only instead of directly accessing the source data to preserve the privacy issue.

In this work, we investigate multi-source-free unsupervised domain adaptation (MSFDA) \cite{Ahmed_2021_CVPR} \cite{NEURIPS2021_Dong} and improve the typical SFDA settings \cite{liang2020we} \cite{yang2022attracting} by introducing multiple source domains.
It therefore holds the potential to serve as an appealing solution for real-world large-scale medical image analysis studies involving multiple centres.
Several recent efforts have been made \cite{NEURIPS2021_Dong} \cite{han2023discriminability} with preliminary attempts to the self-supervised clustering pseudo-labelling method \cite{wang2023mosaic}, which is commonly adopted for MSFDA. 
However, they tend to be suboptimal particularly for medical image processing. 
Since the distinctions of the data from multiple centres are large, the models trained on datasets derived from single or multiple healthcare institutions have not demonstrated a consistent ability to generalise their applicability to external sites \cite{zech2018variable}.

To transcend the aforementioned bottlenecks, in this paper, we propose a framework for MSFDA for medical image analysis. Our contributions include:

1) We propose a novel algorithm termed as \textit{\textbf{U}ncertainty-aware \textbf{A}daptive \textbf{D}istillation} (UAD). 
Our algorithm first recognises the source model with the most comparable underlying data distribution to the target domain to deliver coordinated model initialisation, and then further leverages the complementary knowledge among source models for precise distillation to the target domain;
2) To avoid over- and under-confidence issues, we apply the Temperature Scaling (TS) method for comprehensive confidence calibration over source models towards a well-regulated knowledge distillation procedure;
3) We substantiate the effectiveness of the proposed method by comparison experiments and ablation studies across diverse scenarios,  demonstrating its practical benefits towards various endpoints with clinical significance.




\begin{figure*}[t!]
  \centering
  \centerline{\includegraphics[width=14cm]{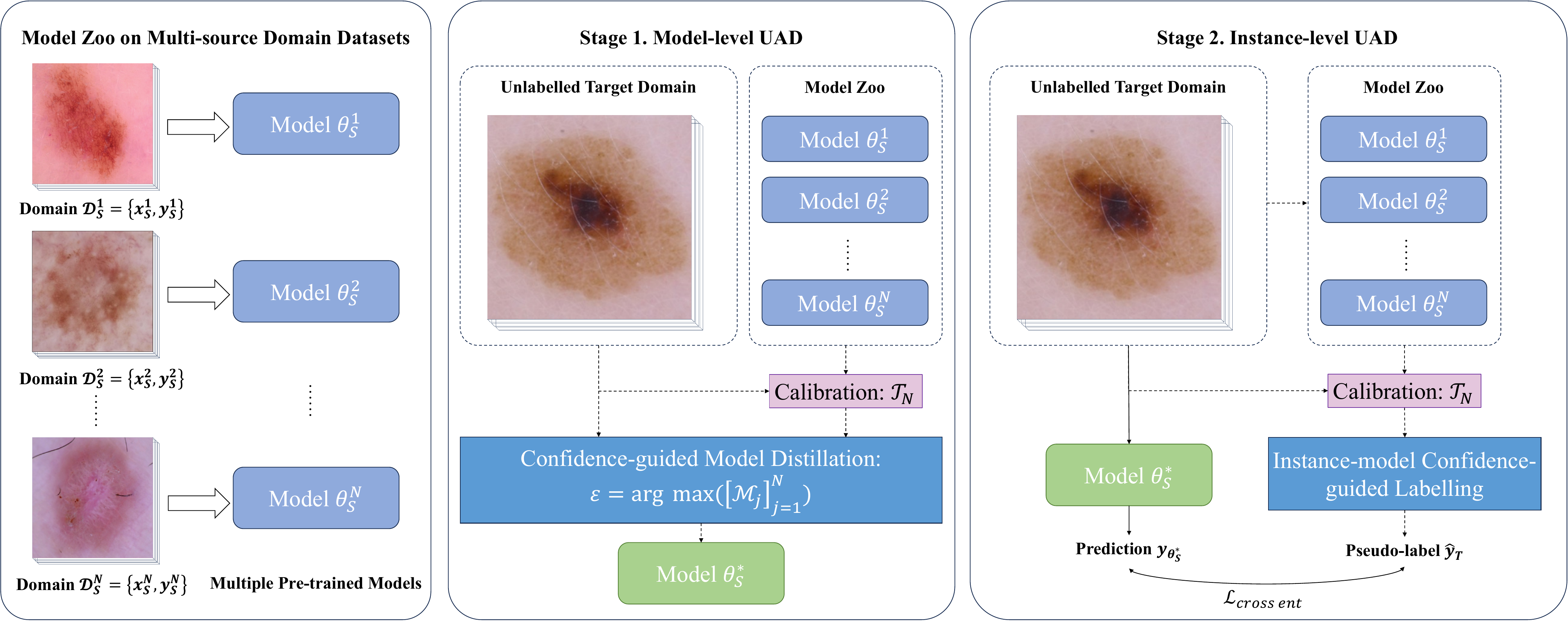}}
\caption{\textbf{Overview of the proposed framework.} Our framework follows a multi-source domain model pre-training process with a two-stage uncertainty-aware adaptive distillation (UAD) process of model initialisation and pseudo-labelling.}
\label{framework}
\vspace{-0.5cm}
\end{figure*}

\section{Methods}
\label{sec:methods}

\subsection{Problem Setting}
Without involving any source domain data in training the final model, we aim to transfer a series of models, pre-trained on multiple source domains, to a new target domain without any human annotation.
In this work, we will consider the $K$-way classification-model adaptation. 
We are given a source model zoo $\{\theta_S^{j}\}_{j=1}^{N}$, which contains $N$ source classification models from $N$ source domains. 
For the $j$-th source model $\theta_S^{j}$ in the source model zoo, with the input space being $\mathcal{X}$ and the output space being $\mathcal{Y}$, it is learned by the source dataset $\mathcal{D}_S^j=\{x_{S_j}^i, y_{S_j}^i\}_{i=1}^{n_j}$ with $n_j$ instances, where $x_{S_j}^{i} \in \mathcal{X}_{S_j}$, $y_{S_j}^{i}\in \mathcal{Y}_{S_j}$. 
A target classification model $\theta_T: \mathcal{X} \rightarrow \mathbb{R}^K$ is learned by only $\{\theta_S^{j}\}_{j=1}^{N}$ and the unlabelled target domain dataset $\mathcal{D}_T=\{x^i_T\}_{i=1}^{n_T}$ with $n_T$ instances.

\subsection{Uncertainty-aware Adaptive Distillation}\label{UAD}

In the proposed framework, we transfer the knowledge from multiple source models to adapt the target domain with pseudo-labels generated by distilling the proper source model. 
Technically, we learn a set of uncertainty (or its opposite, confidence) measures for both overall domain-wise and individual instance-wise distillation corresponding to each source model in the source model zoo.
It evaluates the distributional distance of certain source models working on the target domain dataset and the quality of pseudo-labelling. 
Specifically, we introduce \textit{margin}, defined as the difference between the predicted probabilities of the first and second most probable classes \cite{Settles2009ActiveLL}, as the metric to estimate the confidence measure:
\begin{small}
\begin{align}
\mathcal{M} = \topk_{k=1}(\delta(\theta(x))) - \topk_{k=2}(\delta(\theta(x))),\
\label{margin}
\end{align}
\end{small}
where $\delta(\cdot)$ denotes the Softmax Layer operation with $\delta_j(v)=\frac{\exp(v_j)}{\sum_{i=1}^K \exp(v_i)}$ for $j = 1,...,K$ and $v \in \mathbb{R}^K\mapsto(0,1)^K$. 
Intuitively, if a model $\theta$ has a larger value of the margin $\mathcal{M}$ while predicting an instance, it is regarded as more optimal to extract the instance's feature and finally does the classification task.

In order to prevent the trained target domain model from being interrupted by confounding factors incurred by attributed irrelevant to the target task (e.g., image appearance discrepancy due to inconsistent imaging protocols) or avoid local minima problems, we propose to perform Uncertainty-aware Adaptive Distillation (UAD) from two complementary perspectives, (i) model-level and (ii) instance-level, towards directed and well-regularised multi-source model adaptation. 
The overview of our proposed framework is illustrated in Fig.~\ref{framework}.

\noindent \textbf{\textit{Model-level UAD:}} In previous work related to multi-source domain adaptation\cite{Ahmed_2021_CVPR}, it was a common practice to involve all source models with varying weights in the subsequent fine-tuning stage. 
However, we found that if there is a significant domain gap between a particular source model and the target domain, negative transfer \cite{8953565} could be incurred which results in biased adaptation.
Thus, to initialise a base target model with minimal disturbance, we collect all pre-trained source models from each domain and estimate the overall confidence measure of each source model for predicting the target domain data. 
Specifically, for assessing the confidence of a source model $\theta_S^j$'s inference results on the target domain data, we average all confidence measures estimated for each instance of the target domain data as follows:
$\mathcal{M}_{\text{j}} = \frac{\sum_{i=1}^{n_T} \mathcal{M}_{\text{i}}} {n_T}$.
The source model with the largest confidence measure which is defined as $\varepsilon$ for the target domain, $\theta_S^\ast$, is regarded as the model conforming to the underlying data distribution closest to the target domain and can be considered as the optimal teacher: 
\begin{small}
\begin{align}
\varepsilon = \text{arg} \ {\text{max}}([\mathcal{M}_{\text{j}}]_{j=1}^{N}).\
\label{pick_margin}
\end{align}
\end{small}
We assign the source model $\theta_S^\ast$ as the initial model for SFDA learning on the target data to minimise the gap between the multiple source domains and the target domain.


\noindent \textbf{\textit{Instance-level UAD:}} As the target domain data are not annotated, we propose to use the instance-level UAD method for self-supervised learning on the target data with pseudo labels.
Specifically, we sequentially estimate the confidence measure (margin) of each model in the source model zoo for predicting each instance $x^i_T$, for $i=1,...,n_T$, in the target domain and select the most confident source model to generate the pseudo-label:
\begin{small}
\begin{align}
\varepsilon_i = \text{arg} \ {\text{max}}([\mathcal{M}_{\text{i}}]_{i=1}^{n_T}),\
\label{single_margin}
\end{align}
\end{small}
where $\mathcal{M}_{\text{i}}$ denotes the margin values of source models predicting the target domain instance with:
\begin{small}
\begin{align}
\mathcal{M}_{\text{i}} = \Big[\topk_{k=1}(\delta(\theta_S^j(x^i_T))) - \topk_{k=2}(\delta(\theta_S^j(x^i_T))) \Big]_{i=1,j=1}^{n_T,N}.
\end{align}
\end{small}

For the instance $x^i_T$, the corresponding pseudo-label is obtained by prediction of the source model with $\mathcal{M}_{\text{i}}=\varepsilon_i$, which we define as $\theta_T^i$:
$\hat{y}^i_T = \Big[\theta_T^i(x^i_T) \Big]_{i=1}^{n_T}$.
$\{x^i_T, \hat{y}^i_T\}_{i=1}^{n_T}$ is leveraged to fine-tune the target initial model $\theta_T = \theta_S^\ast$ by minimising the standard cross-entropy loss:
\begin{small}
\begin{equation}
\mathcal{L}_{tar} = -\mathbb{E}_{(x_{T},\hat{y}_{T})\in \mathcal{X}_{T} \times \mathcal{\hat{Y}}_{T}} \sum\nolimits_{k=1}^{K} \mathbbm{1}_{[k=\hat{y}_{T}]} \log \delta_k(\theta_T(x_{T})),
\label{overall_loss}
\end{equation}
\end{small}
where $\mathbbm{1}(\cdot)$ gives value $1$ when the argument is true.

\subsection{Temperature Scaling} \label{ts}

In certain models, domain shift and limited data in source domains may result in over- and under-confidence in predicting target domain data which potentially triggers a mismatch between model prediction accuracy and confidence \cite{NEURIPS2021_f8905bd3}.
In other words, when this phenomenon occurs, the confidence measure $\varepsilon$ will no longer be an optimal measure for improving model prediction accuracy.

To address this problem, we embedded Temperature Scaling (TS) which acts on prediction probabilities to calibrate the logits prior to confidence measurement. 
In our approach, TS is capable of effectively regularising the representation of uncertainty in model predictions, and a more precise and unbiased representation of uncertainty is preferable for the process of knowledge distillation.
The parameter $\mathcal{T}$ is the so-called temperature, which yields softer probability estimates with larger a temperature to alleviate over-confidence in the model. 
For every source model $[\theta_S^{j}]_{j=1}^{N}$, we learn $\mathcal{T}_{\text{j}}$ by setting an initialisation value $\mathcal{T}_\text{initial}$ and applying temperature scaling on the target domain data $\mathcal{D}_T$: $\mathcal{T}_{\text{j}} = \text{TS-Alg}([\theta_S^{j}]_{j=1}^{N}, \mathcal{D}_T)$.
Specifically, the temperature scaling models are tuned by minimising expected calibration error (ECE), a.k.a., calibration gap, which is defined as the difference between accuracy and confidence for a given bin \cite{pmlr-v70-guo17a}:
\begin{small}
\begin{align}
\text{ECE} = \sum_{m=1}^{M}\frac{|B_m|}{n_T}\Big|\text{acc}(B_m) - \text{conf}(B_m)\Big|,\
\label{ece}
\end{align}
\end{small}
where $M$ denotes the number of interval bins that we group predictions, and $B_m$ represents the batch of indices of instances allocated in the interval $I_m = (\frac{m-1}{M},\frac{m}{M}]$.

Given the logit vector $\theta_S^j(x^i_T)$ obtained from each source model, the calibrated probabilities are estimated by the formula:
$z_j = \theta_S^j(x^i_T) / \mathcal{T}_{\text{j}}$,
where $z_j$ is the calibrated pre-softmax output (logits) that will be utilised in Sec. \ref{UAD}.

\section{Experiments and Results}
\label{sec:exp&res}

\subsection{Dataset and Implementation Details}


\textbf{\textit{Datasets:}} We evaluate the proposed multi-source-free domain adaptation framework for classification tasks on two series of datasets:
\begin{itemize}
    \item Multi-centre Diabetic Retinopathy (\textit{DR}) dataset: The multi-centre DR dataset, which measures DR grades (no DR, mild DR, moderate DR, severe DR and proliferative DR), consists of three public datasets (domains) \textit{APTOS 2019} \cite{aptos2019-blindness-detection}, \textit{DDR} \cite{li2019diagnostic}, and \textit{IDRiD} \cite{data3030025} with counts $3660$, $13673$, and $516$ correspondingly. 
\end{itemize}
\begin{itemize}
    \item Skin Cancer MNIST \textit{HAM10000} \cite{tschandl2018ham10000}: To investigate the classification of lesions as benign or malignant in different parts of the human body, we split it into four domains by skin lesion locations which are \textit{back}, \textit{face}, \textit{lower extremity}, and \textit{upper extremity} with counts $2192$, $745$, $2077$ and $1118$ respectively.
\end{itemize}
In our experimental process, we reprocess the data by first resizing into $256\times256$ and cropping into size $224$; then, we assign one domain as the target in turn while considering the others as source domains.


\begin{table*} [t!]
    \setlength{\belowcaptionskip}{1pt}
    \begin{center}
        \caption{\textbf{Comparison experiments with baselines and ablation study.}
	For method, M-UAD, I-UAD and TS are abbreviations of \textit{model-level UAD}, \textit{instance-level UAD} and \textit{temperature scaling}. 
        For datasets, A, D and I are abbreviations of \textit{APTOS 2019}, \textit{DDR} and \textit{IDRiD}; B, F, L and U are abbreviations of \textit{back}, \textit{face}, \textit{lower extremity} and \textit{upper extremity}. 
        The first three rows are baselines, and the last four rows are ablation study.
        All values are adaptation accuracy (\%).
        The last row is our default method setting and corresponding experimental result.}
    \setlength{\tabcolsep}{1.3mm}{
        \begin{small}
        \begin{tabular}{c|cccc|ccccc}
            \toprule[1.2pt]
            \multirow{2}[2]{*}{Method} & \multicolumn{4}{c|}{DR} & \multicolumn{5}{c}{HAM10000} \\
			\cmidrule(lr){2-5} \cmidrule(lr){6-10}
			~ & D, I $\rightarrow$ A & A, I $\rightarrow$ D & A, D $\rightarrow$ I & \cellcolor{gray!25} AVG. & F, L, U $\rightarrow$ B & B, L, U $\rightarrow$ F & B, F, U $\rightarrow$ L & B, F, L $\rightarrow$ U & \cellcolor{gray!25} AVG.\\
			\midrule[1.2pt]
            AaD (22') \cite{yang2022attracting}  & 36.13 & 33.07 & 46.32 & \cellcolor{gray!25} 38.51 & 64.55 & 64.30 & 65.14 & 72.36 & \cellcolor{gray!25} 66.59\\
            DECISION (21') \cite{Ahmed_2021_CVPR} & 57.32 & 45.43 & 58.33 & \cellcolor{gray!25} 53.69 & 74.27 & \textbf{76.24} & 71.06 & 78.98 & \cellcolor{gray!25} 75.14\\
            CAiDA (21') \cite{NEURIPS2021_Dong} & 71.74 & 44.98 & 50.97 & \cellcolor{gray!25} 55.90 & 73.68 & 73.83 & 79.59 & 78.80 & \cellcolor{gray!25} 76.48\\ 

             \midrule
             M-UAD       & 71.49 & 62.03 & 50.39 & \cellcolor{gray!25} 61.30 & 81.84 & 68.19 & 87.48 & 83.27 & \cellcolor{gray!25} 80.20\\ 
             I-UAD       & 72.91 & 63.71 & 53.10 & \cellcolor{gray!25} 63.24 & 84.58 & 69.66 & 88.78 & 83.09 & \cellcolor{gray!25} 81.53\\ 
             M-UAD + I-UAD     & 74.47 & 64.39 & 53.88 & \cellcolor{gray!25} 64.25 & 85.40 & 71.41 & 89.41 & 84.08 & \cellcolor{gray!25} 82.58\\ 
             \textbf{M-UAD + I-UAD + TS} & \textbf{74.52} & \textbf{65.27} & \textbf{58.72} & \cellcolor{gray!25} \textbf{66.17} & \textbf{85.40} & 73.29 & \textbf{89.70} & \textbf{84.44} & \cellcolor{gray!25} \textbf{83.21}\\ 
            \bottomrule[1.2pt]
    \end{tabular}
    \end{small}
    \label{comparison_experiments}}
    \end{center}
\vspace{-0.9cm}
\end{table*}

\noindent \textbf{\textit{Implementation Details:}} 
Following the top-rank solution for medical image classification \cite{liu2021federated}, we employ DenseNet-121 as the backbone. In the source model training process, we use smooth labels instead of the usual one-hot labels to reduce overfitting and label noise. 
The maximum number of epochs $\mathcal{N}_\text{epoch}$ for both DR and HAM10000 datasets is set to $100$; while during the UAD process, the $\mathcal{N}_\text{epoch}$ is set to $15$ with a series of updated pseudo-labels at the start of each. 
The batch size is set to 32.
For each epoch, there are $\mathcal{N}_\text{training data}/32$ iterations in domains.
We use $\mathcal{T}_\text{initial}=\log{(1/1.5)}$ and $1.5$ for the DR dataset and the HAM10000 dataset, respectively.
For both source models pre-training and adaptive distillation, we leverage stochastic gradient descent with momentum value $0.9$ and weight decay $10^{-3}$, with the learning rate scheduling method \cite{pmlr-v37-ganin15} during the model learning progress. 

\subsection{Comparison Experiments}


For experimental comparison, we included one existing SFDA framework \textit{AaD} \cite{yang2022attracting} with multi-source extension and two MSFDA frameworks \textit{DECISION} \cite{Ahmed_2021_CVPR} and \textit{CAiDA} \cite{NEURIPS2021_Dong} as baseline methods. 
We re-implement them following their default settings.
The experimental results are reported in Table~\ref{comparison_experiments}.
The multi-source extension of AaD is implemented via an ensemble that passes the target data through each of the adapted source model
and takes an average of the soft prediction to obtain the test label.
By exploring the experimental results of iterations during the SFDA process for DECISION, we noted that, except for the target domain I in DR and F in HAM10000, the performance of the DECISION model deteriorates as the iterations increase for training the target model. 
This phenomenon is also observed in the CAiDA framework, although the degradation in model performance in the domain adaptation process is not as severe as in the DECISION framework.
Intuitively, in a domain-biased and unsupervised setting, the model overfits to noisy labels when training on the target data.
It is due to the effect of the involvement of inappropriate source models and low-quality pseudo-labels generated.

In comparison with existing frameworks, our proposed method effectively mitigates both factors that could potentially diminish the performance of the target domain model: we identify the most confident source model, excluding inappropriate ones from participating in the training of the target model, and generate the most reliable pseudo-labels through the optimal source model.
The last row in Table~\ref{comparison_experiments} shows that the average accuracy of domain adaptation via UAD (our method) in both datasets significantly outperforms all the baselines.

\subsection{Ablation Study}

Furthermore, we also performed an ablation study on the domain adaptation process: the model-level UAD only without training implementation, the instance-level UAD only without training implementation, and the model-level and instance-level UAD with training but without temperature scaling.

\noindent \textbf{\textit{Effectiveness on Model-level and Instance-level UAD:}} To avoid inappropriate source model(s), which are learned by the source domain data that deviates significantly from the target domain data distribution, from disrupting the final performance of the target domain model, we first propose the exclusion of such disruptive source model(s) during the training process.
Instead, using the model-level UAD (M-UAD) method, we pick the most confident source model, which is also the optimal choice among existing models, to serve as the initialisation of training the target model process.
This establishes a solid foundation in the early stages of model training.
The first row of the ablation study (M-UAD) in Table~\ref{comparison_experiments} demonstrates the result that implementing only M-UAD leads to an improvement of approximately $5\%$ on average compared to the baseline results.

In an unsupervised learning setting, the generation of pseudo-labels is a crucial step in driving the eventual high-performance model.
Instead, the generation of low-quality pseudo-labels leads the target model to gradually fit into these noisy labels, thereby reducing the final performance of the target model.
To prevent this from occurring, we propose using the instance-level UAD (I-UAD) method to identify the most confident label corresponding to an individual instance as its pseudo-label.
The second row of the ablation study (I-UAD) in Table~\ref{comparison_experiments} gives the experimental result that applying the I-UAD method leads to a higher accuracy for the target model compared to the M-UAD approach.

The third row of the ablation study (M-UAD + I-UAD) in Table~\ref{comparison_experiments} gives the experimental result that the performance can be further improved by jointly applying the two-level UAD. 


\noindent \textbf{\textit{Effectiveness on Temperature Scaling:}} According to Sec.~\ref{ts}, to mitigate the problem of over- and under-confidence in certain model(s) predicting the target domain data, TS is an effective method to calibrate the model.
The last row of Table~\ref{comparison_experiments} gives the experimental result of applying the TS approach to our combined UAD framework, showing an improvement in the average accuracy compared to without applying the TS model calibration method.
This effect is particularly pronounced on some target domains with relatively low accuracy, such as domains I and F in the DR and HAM10000 datasets respectively.

\vspace{-0.26cm}
\section{Conclusion}
\label{sec:conclution}
\vspace{-0.08cm}
In this study, we proposed a two-level uncertainty-aware adaptive distillation method termed UAD, a novel deep learning framework for multi-source-free unsupervised domain adaptation on medical imaging data, with successful application on datasets across diseases and human anatomical regions. 
Both initialising the target domain training process by identifying the optimal source model and generating reliable pseudo-labels by leveraging a post-calibrated source model zoo, our method significantly outperforms the existing frameworks performing on the medical imaging data. 
In conclusion, our proposed method can fill the gap in the MSFDA setting in the field of medical image processing and analysis.

{\setstretch{0.86}

\section{Compliance with Ethical Standards}

This research study was conducted retrospectively using human subject data made available in open access by~\cite{aptos2019-blindness-detection,li2019diagnostic,data3030025,tschandl2018ham10000}. Ethical approval was not required as confirmed by the license attached with the open access data.


\bibliographystyle{IEEEbib}

\bibliography{refs}

@article{ben-david_theory_2010,
	title = {A theory of learning from different domains},
	journal = {Machine Learning},
	author = {Ben-David, Shai and Blitzer, John and Crammer, Koby and Kulesza, Alex and Pereira, Fernando and Vaughan, Jennifer Wortman},
	year = {2010},
	pages = {151--175},
}

@InProceedings{pmlr-v37-ganin15,
  title = 	 {Unsupervised Domain Adaptation by Backpropagation},
  author = 	 {Ganin, Yaroslav and Lempitsky, Victor},
  booktitle = 	 {ICML},
  pages = 	 {1180--1189},
  year = 	 {2015},
}

@inproceedings{hsu2020progressive,
  title={Progressive domain adaptation for object detection},
  author={Hsu, Han-Kai and Yao, Chun-Han and Tsai, Yi-Hsuan and Hung, Wei-Chih and Tseng, Hung-Yu and Singh, Maneesh and Yang, Ming-Hsuan},
  booktitle={WACV},
  pages={749--757},
  year={2020}
}

@InProceedings{Ahmed_2021_CVPR,
    author    = {Ahmed, Sk Miraj and Raychaudhuri, Dripta S. and Paul, Sujoy and Oymak, Samet and Roy-Chowdhury, Amit K.},
    title     = {Unsupervised Multi-Source Domain Adaptation Without Access to Source Data},
    booktitle = {CVPR},
    year      = {2021},
    pages     = {10103-10112}
}

@inproceedings{liang2020we, 
 title={Do We Really Need to Access the Source Data? {S}ource Hypothesis Transfer for Unsupervised Domain Adaptation}, 
 author={Liang, Jian and Hu, Dapeng and Feng, Jiashi}, 
 booktitle={ICML},  
 pages={6028--6039},
 year={2020}
}

@article{Settles2009ActiveLL,
  title={Active learning literature survey},
  author={Settles, Burr},
  year={2009},
  journal={University of Wisconsin-Madison Department of Computer Sciences}
}

@article{aptos2019-blindness-detection,
    author = {Karthik, Maggie, Sohier Dane},
    title = {APTOS 2019 Blindness Detection},
    journal={Kaggle},
    year = {2019},
    url = {https://kaggle.com/competitions/aptos2019-blindness-detection}
}

@Article{data3030025,
AUTHOR = {Porwal, Prasanna and Pachade, Samiksha and Kamble, Ravi and Kokare, Manesh and Deshmukh, Girish and Sahasrabuddhe, Vivek and Meriaudeau, Fabrice},
TITLE = {{Indian Diabetic Retinopathy Image Dataset (IDRiD): A Database for Diabetic Retinopathy Screening Research}},
JOURNAL = {Data},
YEAR = {2018},
}

@article{tschandl2018ham10000,
  title={The HAM10000 dataset, a large collection of multi-source dermatoscopic images of common pigmented skin lesions},
  author={Tschandl, Philipp and Rosendahl, Cliff and Kittler, Harald},
  journal={Scientific data},
  pages={1--9},
  year={2018},
}

@article{li2019diagnostic,
  title={Diagnostic assessment of deep learning algorithms for diabetic retinopathy screening},
  author={Li, Tao and Gao, Yingqi and Wang, Kai and Guo, Song and Liu, Hanruo and Kang, Hong},
  journal={Information Sciences},
  pages={511--522},
  year={2019},
}

@InProceedings{pmlr-v70-guo17a,
  title = 	 {On Calibration of Modern Neural Networks},
  author =       {Chuan Guo and Geoff Pleiss and Yu Sun and Kilian Q. Weinberger},
  booktitle = 	 {ICML},
  pages = 	 {1321--1330},
  year = 	 {2017},
}

@inproceedings{NEURIPS2021_f8905bd3,
 author = {Karandikar, Archit and Cain, Nicholas and Tran, Dustin and Lakshminarayanan, Balaji and Shlens, Jonathon and Mozer, Michael C and Roelofs, Becca},
 booktitle = {NeurIPS},
 pages = {29768--29779},
 title = {Soft Calibration Objectives for Neural Networks},
 year = {2021}
}

@inproceedings{wang2022cris,
  title={Cris: Clip-driven referring image segmentation},
  author={Wang, Zhaoqing and Lu, Yu and Li, Qiang and Tao, Xunqiang and Guo, Yandong and Gong, Mingming and Liu, Tongliang},
  booktitle={CVPR},
  pages={11686--11695},
  year={2022}
}

@article{liu2021federated,
  title={Federated Semi-supervised Medical Image Classification via Inter-client Relation Matching},
  author={Liu, Quande and Yang, Hongzheng and Dou, Qi and Heng, Pheng-Ann},
  journal={MICCAI},
  year={2021}
}

@inproceedings{NEURIPS2021_Dong,
 author = {Dong, Jiahua and Fang, Zhen and Liu, Anjin and Sun, Gan and Liu, Tongliang},
 booktitle = {NeurIPS},
 pages = {2848--2860},
 title = {Confident Anchor-Induced Multi-Source Free Domain Adaptation},
 year = {2021}
}

@inproceedings{
yang2022attracting,
title={Attracting and Dispersing: A Simple Approach for Source-free Domain Adaptation},
author={Shiqi Yang and Yaxing Wang and Kai Wang and Shangling Jui and Joost van de Weijer},
booktitle={NeurIPS},
year={2022},
}

@inproceedings{han2023discriminability,
  title={Discriminability and transferability estimation: a Bayesian source importance estimation approach for multi-source-free domain adaptation},
  author={Han, Zhongyi and Zhang, Zhiyan and Wang, Fan and He, Rundong and Su, Wan and Xi, Xiaoming and Yin, Yilong},
  booktitle={AAAI},
  pages={7811--7820},
  year={2023}
}

@article{Liu2022DecomposeTA,
  title={Decompose to Adapt: Cross-Domain Object Detection Via Feature Disentanglement},
  author={Dongnan Liu and Chaoyi Zhang and Yang Song and Heng Huang and Chenyu Wang and Michael H Barnett and Weidong Cai},
  journal={IEEE Transactions on Multimedia},
  year={2022},
  pages={1333-1344},
}

@inproceedings{liu2020unsupervised,
  title={Unsupervised instance segmentation in microscopy images via panoptic domain adaptation and task re-weighting},
  author={Liu, Dongnan and Zhang, Donghao and Song, Yang and Zhang, Fan and O'Donnell, Lauren and Huang, Heng and Chen, Mei and Cai, Weidong},
  booktitle={CVPR},
  pages={4243--4252},
  year={2020}
}

@INPROCEEDINGS{8953565,
  author={Wang, Zirui and Dai, Zihang and Póczos, Barnabás and Carbonell, Jaime},
  booktitle={CVPR}, 
  title={Characterizing and Avoiding Negative Transfer}, 
  year={2019},
  pages={11285-11294},
}

@article{zech2018variable,
  title={Variable generalization performance of a deep learning model to detect pneumonia in chest radiographs: a cross-sectional study},
  author={Zech, John R and Badgeley, Marcus A and Liu, Manway and Costa, Anthony B and Titano, Joseph J and Oermann, Eric Karl},
  journal={PLoS medicine},
  pages={e1002683},
  year={2018},
}

@InProceedings{Fan_2023_ICCV,
    author    = {Fan, Jianan and Liu, Dongnan and Chang, Hang and Huang, Heng and Chen, Mei and Cai, Weidong},
    title     = {Taxonomy Adaptive Cross-Domain Adaptation in Medical Imaging via Optimization Trajectory Distillation},
    booktitle = {ICCV},
    year      = {2023},
    pages     = {21174-21184}
}

@article{fan2024learning,
  title={Learning to Generalize over Subpartitions for Heterogeneity-aware Domain Adaptive Nuclei Segmentation},
  author={Fan, Jianan and Liu, Dongnan and Chang, Hang and Cai, Weidong},
  journal={International Journal of Computer Vision},
  year={2024}
}

@inproceedings{
wang2023mosaic,
title={Mosaic Representation Learning for Self-supervised Visual Pre-training},
author={Zhaoqing Wang and Ziyu Chen and Yaqian Li and Yandong Guo and Jun Yu and Mingming Gong and Tongliang Liu},
booktitle={ICLR},
year={2023}
}

@inproceedings{li2022domain,
  title={Domain adaptive nuclei instance segmentation and classification via category-aware feature alignment and pseudo-labelling},
  author={Li, Canran and Liu, Dongnan and Li, Haoran and Zhang, Zheng and Lu, Guangming and Chang, Xiaojun and Cai, Weidong},
  booktitle={MICCAI},
  pages={715--724},
  year={2022}
}

}
\end{document}